\documentclass[11pt]{article}
\usepackage{eamt23}
\usepackage{times}
\usepackage{url}
\usepackage{latexsym}
\usepackage[small,bf]{caption} 
\usepackage{tabularx}
\usepackage{adjustbox}
\usepackage{booktabs}
\usepackage{multirow}
\usepackage{amssymb}
\usepackage{cuted}
\usepackage{comment}
\usepackage{lipsum}
\usepackage{color}
\usepackage{CJKutf8}
\usepackage{relsize}
\usepackage{amsmath,amsfonts}
\usepackage{algorithmic}
\usepackage{algorithm}
\usepackage{array}
\usepackage{textcomp}
\usepackage{stfloats}
\usepackage{url}
\usepackage{verbatim}
\usepackage{graphicx}

\DeclareMathOperator*{\argmax}{arg\,max}

\title{Variable-length Neural Interlingua Representations for Zero-shot \\ Neural Machine Translation}

\author{
Zhuoyuan Mao $^{1}$ \hspace{0.3em}
Haiyue Song $^1$ \hspace{0.3em} 
{\bf Raj Dabre $^2$} \hspace{0.3em}
{\bf Chenhui Chu $^1$} \hspace{0.3em}
{\bf Sadao Kurohashi $^{1, 3}$}\\
$^1$ Kyoto University, Japan \hspace{1em}
$^2$ NICT, Japan \hspace{1em} 
$^3$ NII, Japan \\
\texttt{\{zhuoyuanmao, song, chu, kuro\}@nlp.ist.i.kyoto-u.ac.jp} \\
\texttt{raj.dabre@nict.go.jp}
}

\date{}

\begin{document}
\maketitle
\begin{abstract}
The language-independency of encoded representations within multilingual neural machine translation (MNMT) models is crucial for their generalization ability on zero-shot translation. Neural interlingua representations have been shown as an effective method for achieving this. However, fixed-length neural interlingua representations introduced in previous work can limit its flexibility and representation ability. In this study, we introduce a novel method to enhance neural interlingua representations by making their length variable, thereby overcoming the constraint of fixed-length neural interlingua representations. Our empirical results on zero-shot translation on OPUS, IWSLT, and Europarl datasets demonstrate stable model convergence and superior zero-shot translation results compared to fixed-length neural interlingua representations. However, our analysis reveals the suboptimal efficacy of our approach in translating from certain source languages, wherein we pinpoint the defective model component in our proposed method.
\end{abstract}

\section{Introduction}
\begin{figure}[t]
    \centering
    \includegraphics[width=\linewidth]{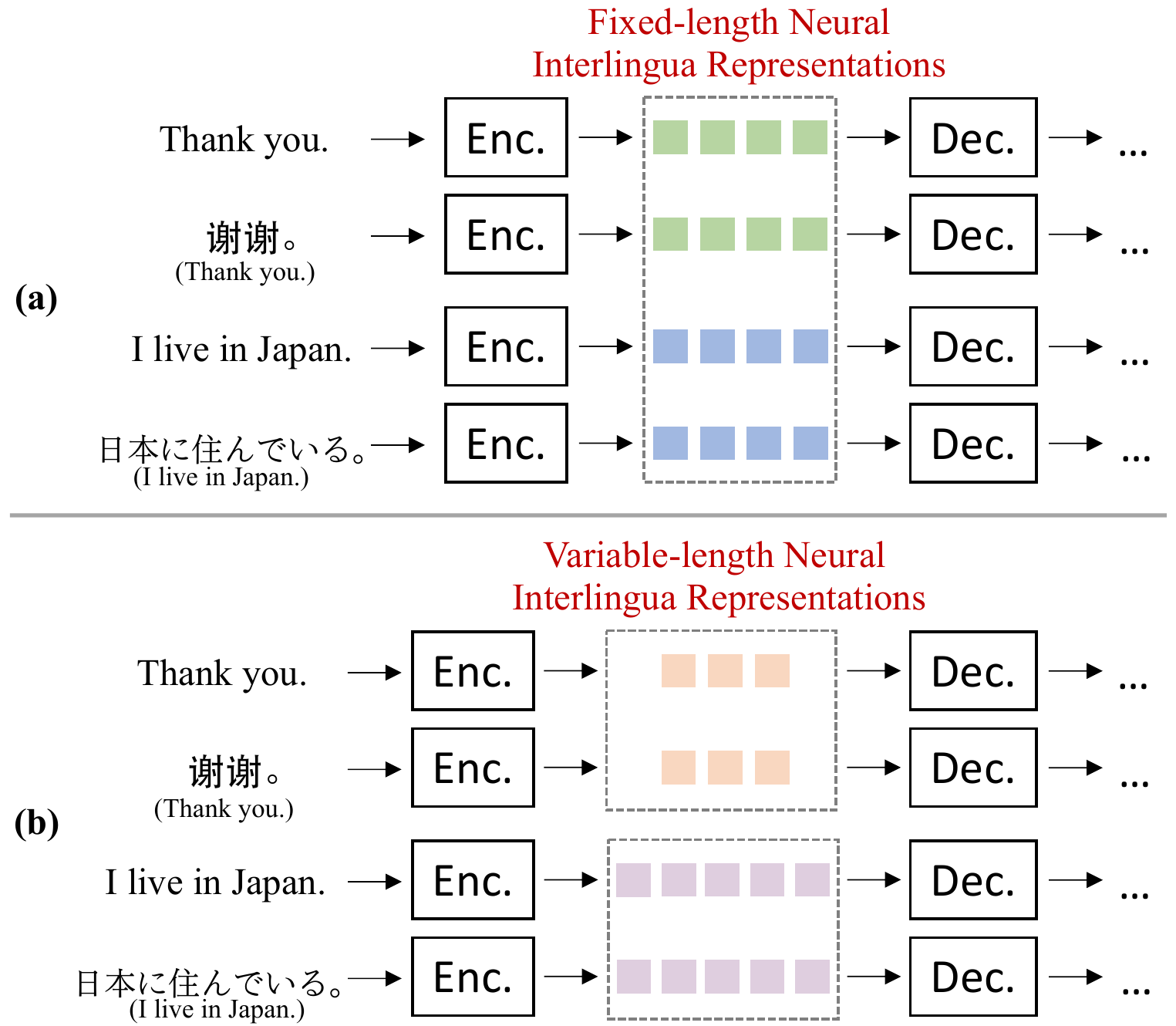}
    \caption{\textbf{(a) Previous fixed-length neural interlingua representations; (b) Our proposed variable-length neural interlingua representations.} Each colored box denotes the representation ($\mathbb{R}^{d\times 1}$) on the corresponding position. ``Enc.'', ``Dec.'', and ``d'' are encoder, decoder, and dimension of model hidden states. }
    \label{fig:intl}
\end{figure}


Multilingual neural machine translation (MNMT)~\cite{dong-etal-2015-multi,firat-etal-2016-multi,ha-etal-2016-toward,johnson-etal-2017-googles,DBLP:journals/csur/DabreCK20} systems enable translation between multiple language pairs within a single model by learning shared representations across different languages. One of the key challenges in building effective MNMT systems is zero-shot translation performance involving unseen language pairs.

Previous work reveals that improving the language-independency of encoded representations is critical for zero-shot translation performance, with neural interlingua representations~\cite{lu-etal-2018-neural,vazquez-etal-2019-multilingual,zhu-etal-2020-language} being proposed as an effective method for achieving this. Neural interlingua representations are shared, language-independent representations that behave as a neural pivot between different natural languages. As shown in Figure~\ref{fig:intl} (a), it enables sentences in different languages with the same meaning to have the same interlingua representations. Previous work has shown the effectiveness of fixed-length neural interlingua representations for zero-shot translation. However, a fixed length can limit neural interlingua representations' flexibility and representation ability. It is highly model size and training data size-sensitive according to our experimental results for different settings of model and training data size.

This paper proposes a novel method for improving neural interlingua representations by making their length variable. As shown in Figure~\ref{fig:intl} (b), our method enables the length of the interlingua representations to vary according to different lengths of source sentences, which may provide more flexible neural interlingua representations. Specifically, we utilize the sentence length in the centric language\footnote{In this work, we consider using an $x$-centric parallel corpus, wherein all sentence pairs within the corpus consist of sentences in language $x$ paired with another language. It is noteworthy that the English-centric corpus is the most prevalent setting. We denote a language distinct from $x$ as a ``non-centric language'' in the subsequent text.} (e.g., English) as the length of neural interlingua representations. We propose a variable-length interlingua module to project sentences in different source languages with the same meaning into an identical neural interlingua representation sequence. To enable translating from non-centric language source sentences during inference, we also introduce a length predictor within the variable-length interlingua module. Moreover, as for the initialization of the interlingua module, we propose a novel method that facilitates knowledge sharing between different interlingua lengths, which can avoid introducing redundant model parameters. We expect that variable-length interlingua representations provide enhanced representations according to different source sentence lengths, which mitigates the model size and training data size-sensitive problem of previous work in low-resource scenarios and improves performance for zero-shot translation.

We conduct experiments on three MNMT datasets, OPUS~\cite{zhang-etal-2020-improving}, IWSLT~\cite{cettolo-etal-2017-overview}, and Europarl~\cite{DBLP:conf/mtsummit/Koehn05} with different settings of training data size and model size. Results demonstrate that our proposed method yields superior results for zero-shot translation compared to previous work. Our method exhibits stable convergence in different settings while previous work~\cite{zhu-etal-2020-language} is highly sensitive to different model and training data sizes. However, we also observe the inferior performance of our method for translation from non-centric language source languages. We attribute it to the accuracy of the interlingua length predictor and point out the possible directions of this research line.



\section{Related Work}
This paper focuses on variable-length interlingua representations for zero-shot NMT.


\subsection{Zero-shot Translation}
In recent years, MNMT~\cite{dong-etal-2015-multi,firat-etal-2016-multi,ha-etal-2016-toward,johnson-etal-2017-googles,aharoni-etal-2019-massively,tan-etal-2019-multilingual,DBLP:journals/csur/DabreCK20,zhang-etal-2020-improving} has been a popular research topic, where the generalization ability of MNMT models to zero-shot translation is a critical problem as obtaining sufficient training data for all translation directions is often impractical. An MNMT model's zero-shot translation performance usually benefits from the encoder-side representations being language-independent and decoder-side representations being language-specific. To achieve this, some studies have proposed removing encoder-side residual connections~\cite{liu-etal-2021-improving-zero} or introducing language-independent constraints~\cite{al-shedivat-parikh-2019-consistency,pham-etal-2019-improving,DBLP:journals/corr/abs-1903-07091,yang-etal-2021-improving-multilingual,DBLP:journals/corr/abs-2305-09312}. Other methods involve decoder pre-training and back-translation~\cite{gu-etal-2019-improved,zhang-etal-2020-improving}, denoising autoencoder objectives~\cite{wang-etal-2021-rethinking-zero}, and encoder-side neural interlingua representations~\cite{lu-etal-2018-neural,vazquez-etal-2019-multilingual,zhu-etal-2020-language}.

\subsection{Neural Interlingua Representations for Zero-shot Translation}
As mentioned above, constructing neural interlingua representations is a powerful method to improve shared encoder representations across various source languages and enhance zero-shot translation. Lu et
al.~\shortcite{lu-etal-2018-neural} first proposed the concept of neural interlingua representations for MNMT, intending to bridge multiple language-specific encoders and decoders using an intermediate interlingua attention module, which has a fixed sequence length. V{\'a}zquez et al.~\shortcite{vazquez-etal-2019-multilingual} extended this approach with a universal encoder and decoder architecture for MNMT and introduced a regularization objective for the interlingua attention similarity matrix. More recently, Zhu et al.~\shortcite{zhu-etal-2020-language} applied the neural interlingua approach in the Transformer~\cite{DBLP:conf/nips/VaswaniSPUJGKP17} model architecture and proposed a position-wise alignment objective to ensure consistent neural interlingua representations across different languages. However, these methods utilized fixed-length neural interlingua representations, which may reduce the model's representation ability for source sentences with different lengths. This paper focuses on revisiting and improving neural interlingua approaches.

\begin{figure*}
    \centering
    \includegraphics[width=\linewidth]{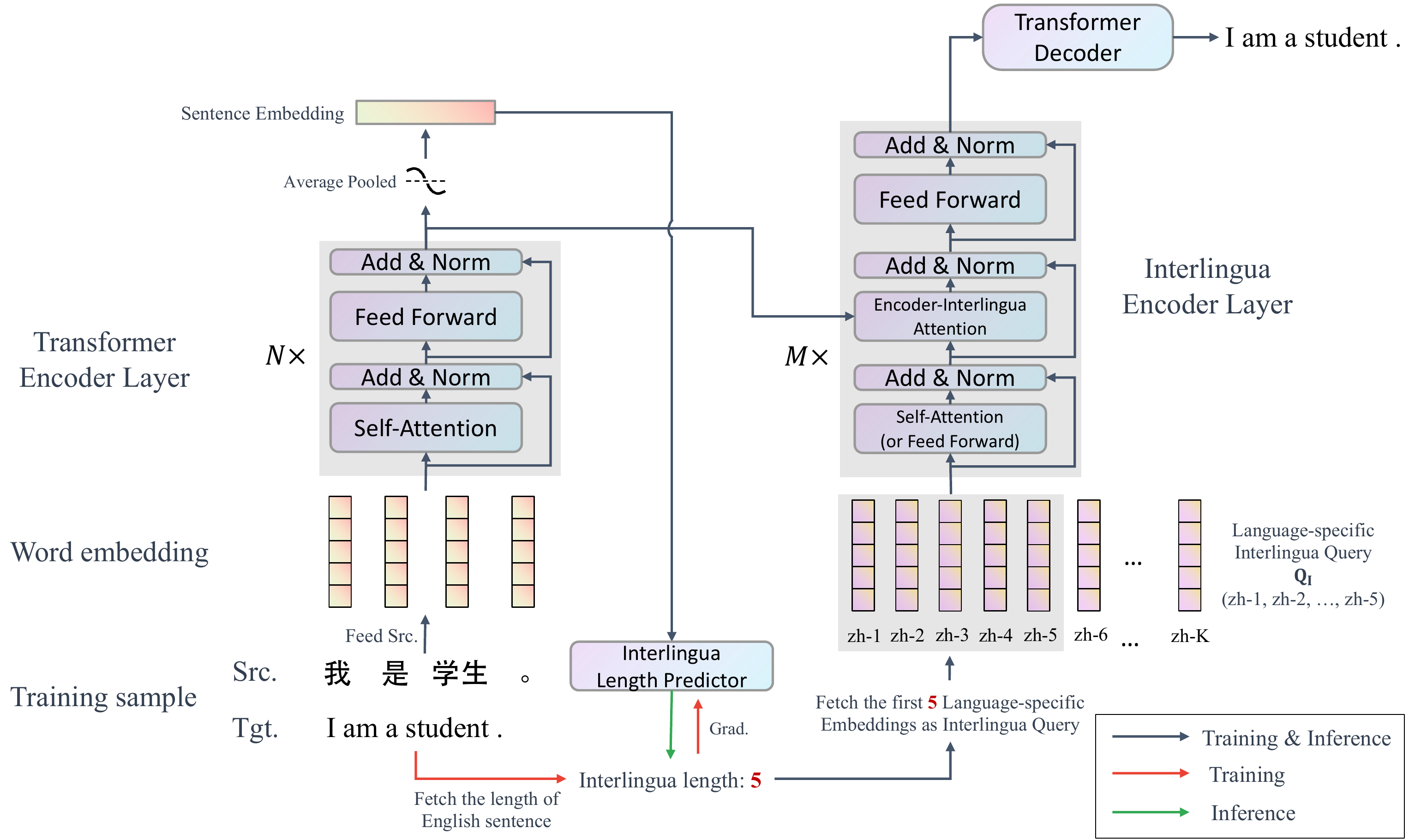}
    \caption{\textbf{Variable-length interlingua module.} ``zh-$x$'' denotes the $x$-th embedding of a Chinese-specific interlingua query.}
    \label{fig:model}
\end{figure*}

\section{Variable-length Neural Interlingua Representations}
\label{sec:method}
We present an MNMT model that comprises three distinct components: a source language encoder, a neural interlingua module, and a decoder. The source language encoder converts source sentences to language-specific representations, the neural interlingua module generates language-agnostic representations, and the decoder converts these representations into the target language translation. In this section, we introduce a novel neural interlingua module.

Specifically, we propose variable-length neural interlingua representations surpassing prior work's fixed-length constraint. To achieve this breakthrough, we have developed a module that includes interlingua encoder layers, an interlingua length predictor, and a language-specific interlingua query. Our module uses an embedding sharing mechanism, as shown in Figure~\ref{fig:model}. Moreover, we introduce the objectives that guide the training of variable-length neural interlingua representations.

\subsection{Variable-length Interlingua Module}
\label{sec:intl}

\noindent{\textbf{Interlingua Encoder Layers}}
In accordance with Zhu et al.~\shortcite{zhu-etal-2020-language}, we construct a variable-length interlingua module within a Transformer model architecture. Our model utilizes $N$ Transformer encoder layers and $6$ Transformer decoder layers, with $M$ interlingua encoder layers introduced between them. To maintain consistency with a standard $6$-layer Transformer encoder, we set $M+N=6$, ensuring that the number of model parameters remains almost the same. Each interlingua encoder layer consists of a sequential series of operations, including self-attention mechanisms (or feed-forward networks),\footnote{We utilize feed-forward networks for the first interlingua encoder layer and employ a self-attention mechanism for subsequent layers. This is because the interlingua query is initially weak and unable to capture similarities through a self-attention mechanism. This design choice is similar to that of Zhu et al.~\shortcite{zhu-etal-2020-language}.} encoder-interlingua attention, and feed-forward networks, as illustrated in Figure~\ref{fig:model}.

The input representations for interlingua encoder layers are denoted as $\mathbf{Q}_{\mathrm{I}}\in \mathbb{R}^{d\times \mathrm{len}_{\mathrm{I}}(X)}$, where $d$ and $\mathrm{len}_{\mathrm{I}}(X)$ respectively indicates the dimension of hidden representations and the length of the neural interlingua representations given a source sentence $X=x_{1}, x_{2}, ..., x_{k}$. Specifically, we define $\mathrm{len}_{\mathrm{I}}(X)$ as follows:
\begin{equation}
    \mathrm{len}_{\mathrm{I}}(X) = \left\{
\begin{aligned}
\mathrm{len}(X) &,& \hspace{0.5em} X \text{is in centric} \\
\mathrm{len}(\mathrm{CT}(X)) &,& \hspace{0.5em} X \text{is in non-centric}
\end{aligned}
\right.,
\label{eq:lenintl}
\end{equation}
where $\mathrm{CT}(X)$ denotes the translation of $X$ in the centric language. We use teacher forcing to generate interlingua length during training. For instance, if we use English-centric parallel sentences as training data, $\mathrm{len}_{\mathrm{I}}(X)$ for each sentence pair will be the length of English sentences. Thus, sentences that convey the same semantic meaning can have the same interlingua length, and interlingua length is variable according to different sentences. 
For the initialization of $\mathbf{Q}_{\mathrm{I}}$, we will provide a detailed explanation of how to generate it later in this section.

Subsequently, $\mathbf{Q}_{\mathrm{I}}$ undergoes self-attention (or feed-forward networks), and we obtain the output $\mathbf{Q}_{\mathrm{I}}^{'}$. Assume that the contextualized representations on top of $N$ Transformer encoder layers are $\mathbf{H}_{\mathrm{S}} \in \mathbb{R}^{d\times k}$. Then we establish an encoder-interlingua attention mechanism:
\begin{equation}
    \mathbf{\mathbf{H}_{\mathrm{EI}}} = \mathrm{Attn}(\mathbf{Q}_{\mathrm{I}}^{'}, \mathbf{H}_{\mathrm{S}}, \mathbf{H}_{\mathrm{S}}),
\end{equation}
where $\mathrm{Attn}(\mathbf{Q}, \mathbf{K}, \mathbf{V})$ indicates the multi-head attention mechanism~\cite{DBLP:conf/nips/VaswaniSPUJGKP17}. This encoder-interlingua attention inherits the design in previous studies of neural interlingua representations~\cite{lu-etal-2018-neural,vazquez-etal-2019-multilingual,zhu-etal-2020-language}.

Finally, we pass $\mathbf{\mathbf{H}_{\mathrm{EI}}}$ through position-wise feed-forward networks to obtain $\mathbf{\mathbf{H}_{\mathrm{I}}}$, the output of the interlingua encoder layers. $\mathbf{\mathbf{H}_{\mathrm{I}}}$ serves as a language-agnostic neural interlingua and can vary in length depending on the source sentence. Once we have $\mathbf{\mathbf{H}_{\mathrm{I}}}$, we feed it into a standard Transformer decoder to generate the translation.

\noindent{\textbf{Interlingua Length Predictor}} Length of interlingua representations is not readily available during inference when translating from non-centric source sentences (e.g., non-English source sentences) using Eq.~\eqref{eq:lenintl}. To address this, we propose using an interlingua length predictor to obtain $\mathrm{len}_{\mathrm{I}}(X)$ for inference. Specifically, we treat the length prediction of translation in the centric language as a classification task, addressed utilizing mean pooled contextualized representations atop the Transformer encoder.\footnote{We attempted to treat it as a regression task, but the performance of the regression model was notably inferior to that of the classifier-based predictor.} More precisely, we predict $X$'s interlingua length as:
\begin{equation}
    \mathrm{len}_{\mathrm{I}}(X) = \argmax_{i}\   \mathrm{softmax}(\frac{\mathbf{1}\ \mathbf{H}_{\mathrm{S}}^{\mathrm{T}}}{k}\mathbf{W}+\mathbf{b})_{i},
    \label{eq:lp}
\end{equation}
where $k$ is the length of $X$, $\mathbf{1}\in \mathbb{R}^{1\times k}$ denotes a vector with all the elements of $1$, $\mathbf{W}\in \mathbb{R}^{d\times K}$ and $\mathbf{b}\in \mathbb{R}^{1\times K}$ indicates the weight and bias of a linear layer, and $K$ is the maximum sequence length allowed in the model.

\noindent{\textbf{Language-specific Interlingua Query}}
Here, we present the method for obtaining input representations $\mathbf{Q}_{\mathrm{I}}$ for the interlingua encoder layers. Initially, we randomly initialize an embedding matrix $\mathbf{E}_{l}\in \mathbb{R}^{d\times K}$ containing $K$ embeddings for the source language $l$. Next, we extract the first $\mathrm{len}_{\mathrm{I}}(X)$ embeddings from $\mathbf{E}_{l}$ to obtain $\mathbf{Q}_{\mathrm{I}}$.
\begin{equation}
    \mathbf{Q}_{\mathrm{I}} = \mathbf{E}_{l} \mathbf{I}_{\mathrm{S}},
    \label{eq:qi}
\end{equation}
where $\mathbf{I}_{\mathrm{S}}\in \mathbb{R}^{K\times \mathrm{len}_{\mathrm{I}}(X)}$ has $1$s as main diagonal elements and $0$s for other elements. Note that the language-specific nature of $\mathbf{E}_{l}$ allows the model to learn a unique mapping from each language to the neural interlingua representations. Zhu et al.~\shortcite{zhu-etal-2020-language} used the technique of language-aware positional embedding~\cite{DBLP:conf/iclr/WangPAN19} for both the neural interlingua representations and the source and target sentences, resulting in ambiguity regarding whether the improvements were from the neural interlingua representations or not. In contrast, our proposed language-specific interlingua query clarifies whether a language-specific mapping to neural interlingua representations benefits zero-shot translation.

\subsection{Training Objectives}
Given a training sample sentence pair $(X,Y)$, we introduce the following training objective, combining an NMT loss, an interlingua alignment loss, and a length prediction loss. The interlingua alignment loss is utilized to guarantee the consistency of the neural interlingua representations for each training sentence pair sample. In contrast, the length prediction loss ensures the generation of variable interlingua length during inference. Specifically, the training objective is defined as follows:
\begin{equation}
    \mathcal{L}(X,Y) = \alpha\mathcal{L}_{\mathrm{NMT}} + \beta\mathcal{L}_{\mathrm{IA}} + \gamma\mathcal{L}_{\mathrm{LP}},
    \label{eq:loss}
\end{equation}
where $\alpha$, $\beta$, and $\gamma$ are weight hyperparameters for each loss, $\mathcal{L}_{\mathrm{LP}}$ is a cross-entropy loss computed from the softmax outputs from Eq.~\eqref{eq:lp}, and $\mathcal{L}_{\mathrm{IA}}$ is a position-wise alignment loss using cosine similarity following Zhu et al.~\shortcite{zhu-etal-2020-language}:
\begin{align}
    \resizebox{\linewidth}{!}{
    $
    \mathcal{L}_{\mathrm{IA}} = 1 - \frac{1}{\mathrm{len}_{\mathrm{I}}(X)}\sum_{i}\cos<\mathbf{H}_{\mathrm{I}}(X)_{i}, \mathbf{H}_{\mathrm{I}}(Y)_{i}>.
    $}
\end{align}
Here $\mathbf{H}_{\mathrm{I}}(\cdot)_{i}$ denotes the $i$-th column of $\mathbf{H}_{\mathrm{I}}(\cdot)$.\footnote{To derive $\mathbf{H}_{\mathrm{I}}(Y)$, it is necessary to feed the target sentence to both the encoder and interlingua encoder layers, which can potentially result in increased computational requirements.} Please note that during training, we always have $\mathrm{len}_{\mathrm{I}}(X)=\mathrm{len}_{\mathrm{I}}(Y)$ because we apply teacher forcing to generate the interlingua length for the sentence pair $(X,Y)$. With $\mathcal{L}_{\mathrm{IA}}$, different sentence pairs with varying lengths of translation in centric language can be represented using variable-length neural interlingua representations. This can enhance the bridging ability for zero-shot translation.

\section{Experimental Settings}

\begin{table}[t]
    \centering
    \resizebox{\linewidth}{!}{
    \begin{tabular}{llrrrrr}
        \toprule
        \textbf{Datasets} & \textbf{Languages} & \textbf{\# Sup.} & \textbf{\# Zero.} & \textbf{\# Train} & \textbf{\# Valid} & \textbf{\# Test} \\
        \toprule
        \multirow{2}{*}{OPUS} & ar, de, en, & \multirow{2}{*}{12} & \multirow{2}{*}{30} & \multirow{2}{*}{12,000,000} & \multirow{2}{*}{2,000} & \multirow{2}{*}{2,000} \\
        & fr, nl, ru, zh & & & & \\
        \hline
        IWSLT & en, it, nl, ro & 6 & 6 & 1,378,794 & 2,562 & 1,147 \\
        \hline
        Europarl & de, en, es, fr, nl & 8 & 12 & 15,782,882 & 2,000 & 2,000 \\
        \bottomrule
    \end{tabular}
    }
    \caption{\textbf{Statistics of the training data}. ``\# Sup.'' and ``\# Zero.'' indicate the respective number of language pairs for supervised and zero-shot translation. ``\# Train'' denotes the total number of the training parallel sentences while ``\# Valid'' and ``\# Test'' showcase the number per language pair.}
    \label{tab:data}
\end{table}

\subsection{Datasets}
Our study involves conducting experiments on zero-shot translation using three distinct datasets, OPUS~\cite{zhang-etal-2020-improving}, IWSLT~\cite{cettolo-etal-2017-overview}, and Europarl~\cite{DBLP:conf/mtsummit/Koehn05}, each comprising $7$, $4$, and $5$ languages, respectively. For 
each dataset, we adopt the train, valid, and test splits following Zhang et al.~\shortcite{zhang-etal-2020-improving}, Wu et al.~\shortcite{wu-etal-2021-language}, and Liu et al.~\shortcite{liu-etal-2021-improving-zero}. Table~\ref{tab:data} presents each dataset's overall statistics. The training and validation data exclusively contains English-centric sentence pairs, indicating the centric language is English in all the experiments, leading to $12$, $6$, and $8$ supervised directions, and $30$, $6$, and $12$ zero-shot directions for each dataset. Refer to Appendix~\ref{app:pre} for preprocessing details.

\subsection{Overall Training and Evaluation Details}
For the OPUS and IWSLT datasets, we utilize a \texttt{Transformer-base} model, while for Europarl, we employ a \texttt{Transformer-big} model, to evaluate the performance of Transformer with both sufficient and insufficient training data. Regarding language tag strategies to indicate the source and target languages to the model, we adopt the method of appending the source language tag to the encoder input and the target language tag to the decoder input~\cite{liu-etal-2020-multilingual-denoising}. This approach allows for the creation of fully language-agnostic neural interlingua representations in between.\footnote{We do not consider employing target language tag appending on the encoder-side~\cite{johnson-etal-2017-googles} in this work because it would require removing both the source and target language information after feeding the source sentence to obtain the neural interlingua representations.} The maximum sentence length is set as $256$, which indicates that $K=256$ (Section~\ref{sec:intl}). Refer to Appendix~\ref{app:train-eval} for other training details.

For evaluation, we choose the evaluation checkpoint based on the validation $\mathcal{L}_{\mathrm{NMT}}$ with the lowest value. We use a beam size of $5$ during inference on the trained models to conduct inference. We report SacreBLEU~\cite{post-2018-call}.\footnote{We utilize the ``zh'' tokenization mode for Chinese, and the ``13a'' tokenization mode for other languages.}

\begin{table*}[t]
    \centering
    \resizebox{\linewidth}{!}{
    \begin{tabular}{l|rrr|rrr|rrr}
        \toprule
        \multirow{2}{*}{Methods} & \multicolumn{3}{c|}{Zero-shot} & \multicolumn{3}{c|}{Supervised: From en} & \multicolumn{3}{c}{Supervised: To en} \\
        & OPUS & IWSLT & Europarl & OPUS & IWSLT & Europarl & OPUS & IWSLT & Europarl \\
        \toprule
        \textit{Pivot} & \textit{22.0} & \textit{19.9} & \textit{29.5} & - & - & - & - & - & - \\
        \hline
        MNMT & 16.5 & 13.1 & 29.0 & \textbf{31.2} & \textbf{29.6} & \textbf{32.9} & \textbf{36.8} & \textbf{33.5} & \textbf{36.1} \\
        Len-fix. Uni. Intl. & 18.2 & 12.7 & 17.4 & 29.6 & 19.6 & 20.1 & 35.3 & 22.2 & 21.8 \\
        Len-fix. LS. Intl. & 18.4 & 4.7 & 5.8 & 30.1 & 7.3 & 6.7 & 35.7 & 12.9 & 7.1 \\
        Len-vari. Intl. (ours) & \textbf{18.9}$^{\dag}$ & \textbf{14.8} & \textbf{29.6} & 30.2$^{\dag}$ & 26.2 & 32.6 & 34.0 & 27.1 & 33.8 \\
        \bottomrule
    \end{tabular}
    }
    \caption{\textbf{Overall BLEU results on OPUS, IWSLT, and Europarl}. The best result among all the settings except \textit{Pivot }is in \textbf{bold}. We mark the results significantly~\cite{koehn-2004-statistical} better than ``Len-fix. Uni. Intl.'' with $\dag$ for OPUS dataset.}
    \label{tab:overall}
\end{table*}

\begin{table*}[t]
    \centering
    \resizebox{\linewidth}{!}{
    \begin{tabular}{l|rrrrrrrrrrrr|r}
        \toprule
        \multirow{2}{*}{Methods} & \multicolumn{2}{c}{de--fr} & \multicolumn{2}{c}{ru--fr} & \multicolumn{2}{c}{nl--de} & \multicolumn{2}{c}{zh--ru} & \multicolumn{2}{c}{zh--ar} & \multicolumn{2}{c|}{nl--ar} & Zero-shot \\
        & $\rightarrow$ & $\leftarrow$ & $\rightarrow$ & $\leftarrow$ & $\rightarrow$ & $\leftarrow$ & $\rightarrow$ & $\leftarrow$ & $\rightarrow$ & $\leftarrow$ & $\rightarrow$ & $\leftarrow$ & Avg. \\
        \toprule
        \textit{Pivot} & \textit{23.4} & \textit{21.2} & \textit{31.0} & \textit{26.0} & \textit{21.8} & \textit{23.6} & \textit{24.8} & \textit{37.9} & \textit{24.0} & \textit{38.9} & \textit{7.4} & \textit{17.4} & \textit{22.0} \\
        \hline
        MNMT & 17.6 & 15.0 & 21.5 & 17.7 & 17.9 & 21.4 & 15.3 & 27.6 & 18.0 & 28.6 & 5.3 & 13.3 & 16.5 \\
        Len-fix. Uni. Intl. & 20.1 & 17.0 & 25.0 & 22.4 & 19.5 & 21.3 & 20.3 & 30.9 & 19.6 & 30.4 & 6.1 & 14.4 & 18.2 \\
        Len-fix. LS. Intl. & \textbf{20.7} & 17.7 & 25.7 & 21.7 & 19.8 & 21.6 & 19.9 & 31.5 & \textbf{20.1} & 31.6 & \textbf{6.5} & \textbf{14.5} & 18.4 \\
        Len-vari. Intl. (ours) & 20.6$^{\dag}$ & \textbf{18.3}$^{\dag}$ & \textbf{26.0}$^{\dag}$ & \textbf{23.4}$^{\dag}$ & \textbf{20.2}$^{\dag}$ & \textbf{22.1}$^{\dag}$ & \textbf{20.8} & \textbf{31.8}$^{\dag}$ & 20.0 & \textbf{31.9}$^{\dag}$ & 6.3 & \textbf{14.5} & \textbf{18.9}$^{\dag}$ \\
        \bottomrule
    \end{tabular}
    }
    \caption{\textbf{BLEU results of zero-shot translation on OPUS}. We randomly select six zero-shot language pairs and report the results. The best result among all the settings except ``\textit{Pivot}'' is in \textbf{bold}. We mark the results significantly~\cite{koehn-2004-statistical} better than ``Len-fix. Uni. Intl.'' with $\dag$.}
    \label{tab:opuszero}
\end{table*}

\subsection{Baselines and Respective Training Details}
To compare our variable-length neural interlingua method with previous fixed-length neural interlingua methods, we trained the following settings:

\noindent \textbf{MNMT}~\cite{johnson-etal-2017-googles} is a system trained with standard \texttt{Transformer-base} or \texttt{Transformer-big} for multiple language pairs. We applied the language tag strategy of source language tag for encoder input and target language tag for decoder input.

\noindent \textbf{Pivot} 
translation~\cite{zoph-knight-2016-multi} involves translating a source language into a pivot language, usually English, and then translating the pivot language into the target language. This system constitutes a robust baseline for zero-shot translation, which we include for reference. We implement this setting by feeding the pivot language output of the MNMT model to itself to generate the target language.

\noindent \textbf{Len-fix. Uni. Intl.}
We follow the setting described by Zhu et al.~\shortcite{zhu-etal-2020-language}, but we remove its language-aware positional embedding to test whether a single interlingua module can improve zero-shot translation. Compared to our variable-length interlingua representations presented in Section~\ref{sec:intl}, these fixed interlingua representations have a universal $\mathrm{len}_{\mathrm{I}}$ (Eq.~\eqref{eq:lenintl}) for different source sentences and a universal $\mathbf{E}\in \mathbb{R}^{d\times \mathrm{len}_{\mathrm{I}}}$ for different languages and without a $\mathbf{Q}_{\mathrm{I}}$ (Eq.~\eqref{eq:qi}). The fixed interlingua length is set to $17$, $21$, and $30$, which are the average lengths of each dataset following Zhu et al.~\shortcite{zhu-etal-2020-language} and V{\'a}zquez et al.~\shortcite{vazquez-etal-2019-multilingual}.

\noindent \textbf{Len-fix. LS. Intl.}
The only difference between this system and the ``Len-fix. Uni. Intl.'' system mentioned above is the initialization of the interlingua query. We use a language-specific $\mathbf{E}_{l}\in \mathbb{R}^{d\times \mathrm{len}_{\mathrm{I}}}$ for each source language $l$ without a $\mathbf{Q}_{\mathrm{I}}$ (Eq.~\eqref{eq:qi}).

\noindent \textbf{Len-vari. Intl. (ours)}
This refers to variable-length neural interlingua  representations proposed in Section~\ref{sec:method}.

For the last three neural interlingua settings, we set $M$ and $N$ to $3$ for both the Transformer encoder and interlingua encoder layers. The values of $\alpha$, $\beta$, and $\gamma$ (Eq.~\eqref{eq:loss}) are set as $1.0$, $1.0$, and $0.1$, respectively. We remove the first residual connection within the first interlingua encoder layer to improve the language-independency of the interlingua representations, inspired by Liu et al.~\shortcite{liu-etal-2021-improving-zero}.

\section{Results and Analysis}
We now present in tables~\ref{tab:overall},~\ref{tab:opuszero}, and~\ref{tab:opussup} the results of our variable-length interlingua approach and compare them against several baselines.

\begin{table*}[t]
    \centering
    \resizebox{\linewidth}{!}{
    \begin{tabular}{l|rrrrrrrrrrrr|rr}
        \toprule
        \multirow{2}{*}{Methods} & \multicolumn{2}{c}{en--ar} & \multicolumn{2}{c}{en--de} & \multicolumn{2}{c}{en--fr} & \multicolumn{2}{c}{en--nl} & \multicolumn{2}{c}{en--ru} & \multicolumn{2}{c|}{en--zh} & \multicolumn{2}{c}{Supervised Avg.} \\
        & $\rightarrow$ & $\leftarrow$ & $\rightarrow$ & $\leftarrow$ & $\rightarrow$ & $\leftarrow$ & $\rightarrow$ & $\leftarrow$ & $\rightarrow$ & $\leftarrow$ & $\rightarrow$ & $\leftarrow$ & From en & To en \\
        \toprule
        MNMT & \textbf{23.9} & \textbf{37.8} & \textbf{30.8} & \textbf{34.6} & \textbf{33.9} & \textbf{35.5} & \textbf{27.8} & \textbf{31.5} & 29.4 & \textbf{35.1} & \textbf{41.2} & \textbf{46.4} & \textbf{31.2} & \textbf{36.8} \\
        Len-fix. Uni. Intl. & 22.6 & 36.6 & 28.9 & 33.0 & 31.7 & 33.5 & 27.4 & 30.1 & 28.4 & 34.0 & 38.8 & 44.6 & 29.6 & 35.3 \\
        Len-fix. LS. Intl. & 22.9 & 36.8 & 29.0 & 33.8 & 32.3 & 33.9 & 27.7 & 30.6 & 28.9 & 34.3 & 39.5 & 44.8 &  30.1 & 35.7 \\
        Len-vari. Intl. (ours) & 23.3$^{\dag}$ & 33.8 & 30.1$^{\dag}$ & 32.3 & 32.9$^{\dag}$ & 32.6 & 27.3 & 27.9 & \textbf{29.5}$^{\dag}$ & 32.2 & 38.0 & 45.3$^{\dag}$ & 30.2$^{\dag}$ & 34.0 \\
        \bottomrule
    \end{tabular}
    }
    \caption{\textbf{BLEU results of supervised translation on OPUS}. The best result among all the settings is in \textbf{bold}. We mark the results significantly~\cite{koehn-2004-statistical} better than ``Len-fix. Uni. Intl.'' with $\dag$.}
    \label{tab:opussup}
\end{table*}

\begin{table*}[t]
    \centering
    \begin{tabular}{lrrrrrrr}
        \toprule
         & ar & de & fr & nl & ru & zh & Avg. \\
        \toprule
        Acc. of Len. Pre. & 20.6 & 26.5 & 17.6 & 19.3 & 21.1 & 13.8 & 19.8 \\
        Avg. of $|$ Len. Pre. $-$ \textit{gold} $|$ & 2.4 & 3.4 & 3.8 & 3.1 & 3.3 & 3.9 & 3.3 \\
        \hline
        BLEU w/ Len. Pre. & 33.8 & 32.3 & 32.6 & 27.9 & 32.2 & 45.3 & 34.0 \\
        BLEU w/ \textit{gold} & 35.5$^\dag$ & 33.4$^\dag$ & 33.3$^\dag$ & 29.4$^\dag$ & 33.4$^\dag$ & 46.0$^\dag$ & 35.2$^\dag$ \\
        \bottomrule
    \end{tabular}
    \caption{\textbf{Accuracy of the interlingua length predictor, averaged absolute difference between predicted length and \textit{gold} length, and ``to en'' BLEU scores of each non-English source language on OPUS.} ``w/ Len. Pre.'' and ``w/ \textit{gold}'' indicate using the predicted interlingua length and the correct interlingua length (length of the English translation), respectively. Accuracy of the length predictor and average abosulute difference are evaluated using OPUS's test set. We mark the results significantly~\cite{koehn-2004-statistical} better than ``BLEU w/ Len. Pre.'' with $\dag$.}
    \label{tab:len}
\end{table*}

\subsection{Main results}
Firstly, Tables~\ref{tab:overall} and~\ref{tab:opuszero} indicate that our proposed variable-length interlingua representations outperform previous work in zero-shot directions. The severe overfitting issue of ``Len-fix. Uni. Intl.'' and ``Len-fix. LS. Intl.'' on IWSLT and Europarl suggests that they are limited to model size and training data size settings, while our proposed method can converge stably on all three settings. These results demonstrate that our flexible interlingua length can benefit zero-shot translation more effectively. Secondly, our proposed method performs better than previous work in ``from en'' supervised directions as shown in Tables~\ref{tab:overall} and~\ref{tab:opussup}, but still falls short of the MNMT baseline. This may be attributed to the interlingua module's weak source-target awareness. Thirdly, our variable-length neural interlingua representations perform significantly worse on ``to en'' directions than ``Len-fix.'' methods on OPUS and MNMT on all datasets. We provide analysis of this phenomenon next.

\subsection{Validation NMT Loss}

\begin{figure}[t]
    \centering
    \includegraphics[width=\linewidth]{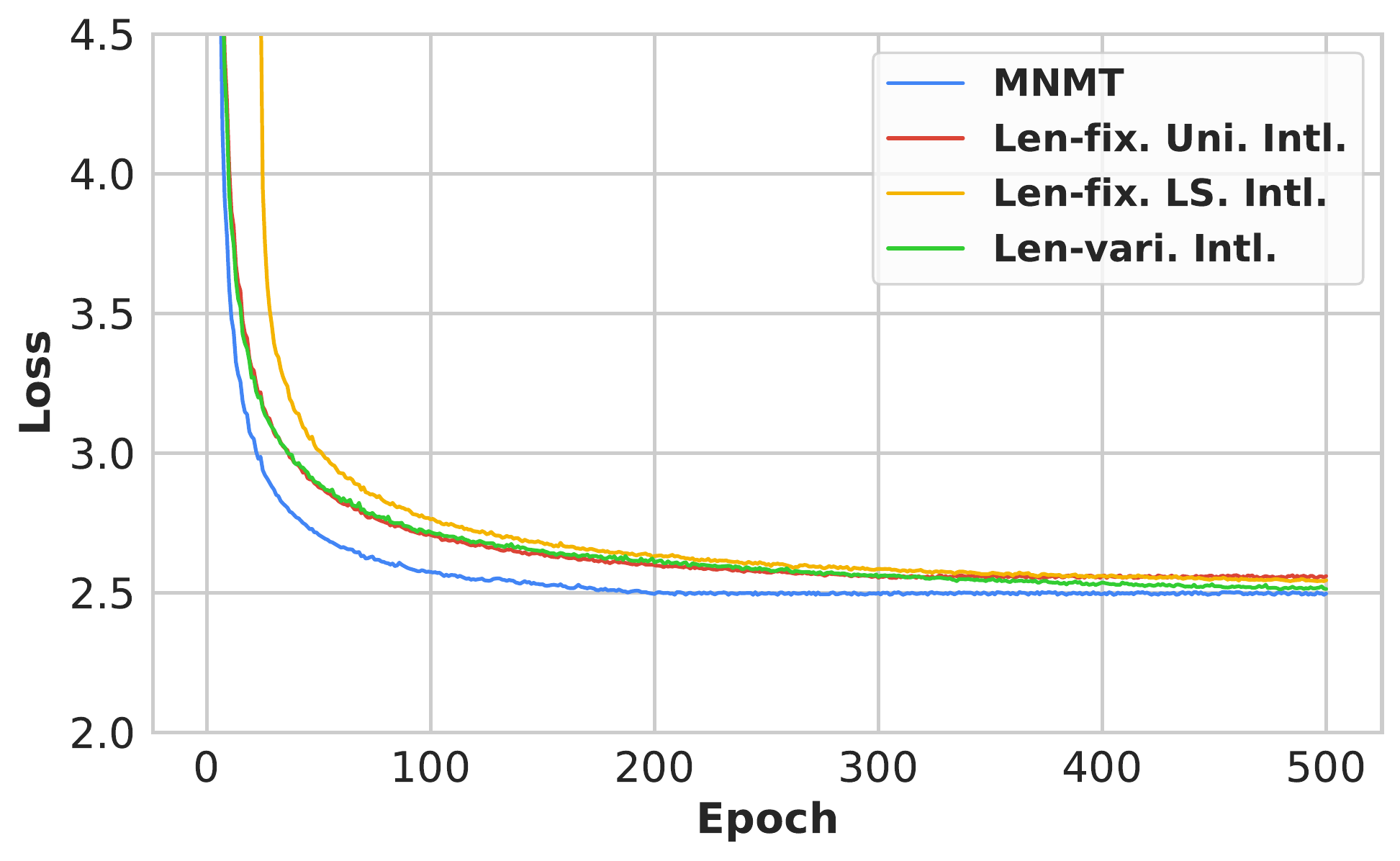}
    \caption{Validation NMT loss curve on OPUS.}
    \label{fig:valid}
\end{figure}

We investigate why variable-length neural interlingua representations perform poorly in ``to en'' supervised directions by analyzing the validation NMT loss, an approximate measure of NMT performance on the validation set. Figure~\ref{fig:valid} displays the validation NMT loss for all settings on OPUS. We observe that variable-length interlingua representations can converge well, even smaller than the validation loss of ``Len-fix. Uni. Intl.'' and ``Len-fix. LS. Intl.'' However, the interlingua length predictor was teacher-forced during training, indicating the validation NMT loss was calculated with a 100\% accurate interlingua length predictor. As a result, the inaccurate interlingua length predictor is likely the primary cause of our method's inferior performance in ``to en'' directions, despite its well-converged validation NMT loss.

\subsection{Impact of the Interlingua Length Predictor}

We analyze the interlingua length predictor and identify the reason for the subpar performance in ``to en'' translations. We input the source sentences of the test set in non-English languages into the model and check whether the predicted length in interlingua is identical to the length of its English reference. We present the accuracy on the OPUS dataset in Table~\ref{tab:len}. The results show that the accuracy for each language is approximately 20.0\%, which can result in error propagation when translating from those languages. To further understand the impact of the length predictor quality on translation performance, we attempt to provide the model with the correct interlingua length instead of relying on the length predictor. As shown in Table~\ref{tab:len}, the results reveal significant BLEU improvements when the correct interlingua length is applied. This suggests that the performance issue encountered when translating from a non-centric source language can be addressed by upgrading the interlingua length predictor's accuracy. Furthermore, we can also enhance zero-shot translation performance if we have a better length predictor. Nevertheless, we observe that even with a low length prediction accuracy of approximately 20.0\%, we can still achieve solid BLEU performance, averaging 34.0 BLEU points. This indicates that an incorrectly predicted length with just a trivial difference, as shown in Table~\ref{tab:len}, will not result in the enormous information loss required for translation.

\section{Conclusion}
This study introduced a novel variable-length neural interlingua approach that improved zero-shot translation results while providing a more stable model than previous fixed-length interlingua methods. Although our analysis revealed a performance downgrade in ``to en'' directions, we have identified the problematic model component and plan to address it in future studies.

\section*{Acknowledgements}
This work was supported by JSPS KAKENHI Grant Number 22KJ1843.

\bibliography{eamt23,anthology}
\bibliographystyle{eamt23}

\appendix
\section{Preprocessing Details}
\label{app:pre}
Jieba\footnote{\url{https://github.com/fxsjy/jieba}} is used to segment Chinese while Moses\footnote{\url{https://github.com/moses-smt/mosesdecoder}}~\cite{koehn-etal-2007-moses} is utilized to tokenize other languages. We employ BPE~\cite{sennrich-etal-2016-neural} with $50,000$, $40,000$, and $50,000$ merge operations to create a joint vocabulary for each dataset, resulting in the vocabulary sizes of $66,158$, $40,100$, and $50,363$, respectively.

\section{Training Details}
\label{app:train-eval}
Our models are trained using Fairseq.\footnote{\url{https://github.com/facebookresearch/fairseq}} As the data size for each language pair is relatively similar, oversampling is not implemented for MNMT. The dropout rate was set to $0.1$, $0.4$, and $0.3$ for each dataset, and we use the Adam optimizer~\cite{DBLP:journals/corr/KingmaB14} with a learning rate of $5$e-$4$, $1$e-$3$, and $5$e-$4$, respectively, employing $4,000$ warm-up steps. 
The \texttt{Transformer-base} model was trained using four 32 GB V100 GPUs, and the \texttt{Transformer-big} model was trained using eight 32 GB V100 GPUs, with a batch size of $4,096$ tokens. To speed up training, mixed precision training~\cite{DBLP:conf/iclr/MicikeviciusNAD18} is also employed. Each dataset is trained for $500$, $200$, and $500$ epochs.

\section{Limitations}
While this study proposed a novel method for improving neural interlingua representations for zero-shot translation, the following limitations should be addressed in future work:
\begin{itemize}
    \item The inaccurate interlingua length predictor currently leads to inferior performance for translation from non-centric languages. Therefore, a better predictor should be explored to improve the performance.
    \item We used the length of centric language sentences as the interlingua length, which may limit the application for using parallel sentences not involving the centric language. Therefore, a better way to generate variable lengths for neural interlingua representations should be developed in future work.
    \item We have yet to test whether the neural interlingua representations obtained in this study can act as a semantic pivot among all the languages. Thus, it would be interesting to evaluate the effectiveness of our variable-length interlingua representations on cross-lingual language understanding tasks~\cite{DBLP:conf/icml/HuRSNFJ20}.
\end{itemize}

\end{document}